\pdfoutput=1

\documentclass[11pt]{article}

\usepackage{acl}

\usepackage{caption}
\usepackage{subcaption}
\usepackage{times}
\usepackage{latexsym}
\usepackage{amsmath}
\usepackage{paralist}
\usepackage{graphicx}
\usepackage{booktabs}
\usepackage{amssymb}
\usepackage{pifont}
\newcommand{\cmark}{\ding{51}}%
\newcommand{\xmark}{\ding{55}}%
\usepackage[T1]{fontenc}

\usepackage[utf8]{inputenc}

\usepackage{microtype}

\usepackage{inconsolata}

\usepackage{placeins}
\usepackage{csquotes}

\title{Approximate Attributions for Off-the-Shelf Siamese Transformers}

\author{Lucas Möller\(^1\) \quad Dmitry Nikolaev\(^2\)\thanks{~~The work was done while Dmitry Nikolaev was a postdoc at the Institute for Natural Language Processing, University of Stuttgart.} \quad Sebastian Padó\(^1\)\\
        \(^1\)Institute for Natural Language Processing, University of Stuttgart, Germany \\ 
        \(^2\)University of Manchester, UK \\ 
        \texttt{\{lucas.moeller, pado\}@ims.uni-stuttgart.de}, \texttt{dmitry.nikolaev@manchester.ac.uk}}

\begin{document}
\maketitle
\begin{abstract}
  Siamese encoders such as sentence transformers are among the least
  understood deep models.
  Established attribution methods cannot tackle this model class since 
  it compares two inputs rather than
  processing a single one. To address this gap, we have recently proposed an 
  attribution method specifically for Siamese encoders \citep{emnlp}.  
  However, it requires models to be adjusted and fine-tuned and therefore 
  cannot be directly applied to off-the-shelf models. 
  In this work, we reassess these restrictions and propose 
  (i)~a model with exact attribution ability that retains the original model's predictive performance and 
  (ii)~a way to compute approximate attributions for off-the-shelf models.
  We extensively compare approximate and exact attributions and use them to analyze the models' attendance to different linguistic aspects.  
  We gain insights into which syntactic roles Siamese transformers attend to, confirm that they mostly ignore negation, explore how they judge semantically opposite adjectives, and find that they exhibit lexical bias.
\end{abstract}

\section{Introduction}
Siamese Encoders (SE) are a class of deep-learning architectures that are trained by
comparing embeddings of two inputs produced by the same encoder.
In NLP
they are often realized in the form of sentence transformers or STs \cite{sbert}, which 
have been successfully applied to the prediction of semantic similarity \cite{sts}, natural language inference \cite{conneau}, and in information retrieval \cite{beir}.

Despite their wide use, our understanding of which aspects of inputs STs base their decisions on is still limited, partly due to the fact that 
established attribution methods like integrated gradients \cite{intgrads} cannot be directly applied to SEs as they compare two inputs rather than processing a single one.

In a recent publication \cite{emnlp}, we have derived an attribution method specifically targeted for SEs by generalizing the concept of integrated gradients to models with two inputs and introduced \textit{integrated Jacobians} (IJ).
Resulting attributions take the form of token--token matrices (cf.\ Figure~\ref{fig:attr_agreement}) and they inherit theoretical guarantees from integrated gradients.
However, they require models to be adjusted in two ways: (1)~embeddings need to be shifted by a reference input and (2)~the usual cosine similarity is replaced by a dot product. This has a number of disadvantages: the (unnormalized) dot-product is not a sufficient similarity measure, the adjustments lead to a drop in predictive performance, and models  need to be fine-tuned.

In this work, we address these drawbacks.
Our main contributions are twofold:

\begin{itemize}
    \setlength\itemsep{0.1em}
    \item We show that it is possible to compute attributions for models using cosine similarity as a similarity measure. A resulting model with \text{exact} attribution ability can retain the downstream performance of the original ST.
    \item  We propose a method to compute \textit{approximate} attributions for off-the-shelf SE models that do not require adjustments. These attributions do not come with the theoretical guarantees of their exact counterparts: They agree with them partly but have their limits.
\end{itemize}
These updates to our original method close the performance gap between standard and interpretable STs. 
Our additional evaluations provide important guidance for the use and the limitations of approximate attributions for off-the-shelf models.
Our code is  available on github at\\ \url{https://github.com/lucasmllr/xsbert}.
%

\section{Related Work}

\paragraph{Model Explainability.}
A large number of concepts and methods are associated with model explainability, and no unified definition exists \cite{murdoch}.
Feature-attribution methods, showing which parts of an input the model consults for a given prediction,
are a means of \textit{local} explainability for individual predictions \cite{li}. 
They provide post-hoc explanations for models that are not inherently interpretable, because we cannot decompose their decision making process into intuitively understandable pieces at prediction time \cite{rudin}.
The framework of Integrated Gradients \citep[IG;][]{intgrads} provides a way to do this in a provably correct way and with measurable accuracy.
In the terminology introduced by \citet{doshi}, such feature attributions are individual \textit{cognitive chunks} that may be cumulated across input dimensions and add up to the total prediction.

\paragraph{Analysis of Transformers.}
A number of publications have analyzed Transformer-based language models \citep{rogers}.
A~lot of attention has been directed towards interpreting the self-attention weights and  visualizing the process of token prediction \citep{clark,voita}.
It has been pointed out, however, that attention weights alone are insufficient for
explaining model predictions \citep{wiegreffe,kobayashi2023analyzing}, 
and \citet{bastings}
conclude that feature attribution methods should be used instead.
The latter were surveyed by \citet{xnlp}, and \citet{atanasova} found IG
to be among the most robust methods.

\paragraph{Analysis of Siamese Transformers.}
Less work aims at better understanding
STs. \citet{opitz} fine-tune STs to encode well-defined AMR-based
semantic features in selected dimensions of the model's embedding space.
\citet{macavaney} focus on IR models and analyze predictions for pairs
of input queries and documents with certain known properties.
\citet{nikolaev-pado-2023-representation} construct synthetic sentence
pairs with specific lexical and syntactic characteristics and regress
similarity scores on these features.  Finally, \citet{emnlp} extend IG
to apply to STs and, as a case study, analyze which parts of speech
STs preferentially attend to (cf. Section~\ref{sec:exact_attr} of this paper).


\section{Method}
\subsection{Exact Attributions}
\label{sec:exact_attr} 

In \citet{emnlp}, we derived an exact attribution method for a Siamese
model $f$ with an encoder $\mathbf{e}$ mapping two inputs $\mathbf{a}$ and $\mathbf{b}$ to a
scalar score $s$:
\begin{equation} \label{eq:biencoder}
    f(\mathbf{a}, \mathbf{b}) = \mathbf{e}^T(\mathbf{a}) \, \mathbf{e}(\mathbf{b}) = s
\end{equation}
Due to space limits, we can only summarize the most important results here; see the original publication for a full derivation. The approach begins by  extending the concept of integrated gradients \citep{intgrads} to the Siamese case:
\begin{equation} \label{eq:two_inputs}
\begin{split}
    & f(\mathbf{a}, \mathbf{b}) - f(\mathbf{a}, \mathbf{r}_b) - f(\mathbf{b}, \mathbf{r}_a) + f(\mathbf{r}_a, \mathbf{r}_b) \\[1.5ex]
    = & \int_{\mathbf{r}_b}^\mathbf{b}\! \int_{\mathbf{r}_a}^\mathbf{a} \frac{\partial^2}{\partial \mathbf{x}_i \partial \mathbf{y}_j} \, f \left(\mathbf{x}, \mathbf{y}\right) \,d\mathbf{x}_i \,d\mathbf{y}_j \\[1ex]
    = & \sum_{ij} \left(\mathbf{a} - \mathbf{r}_a \right)_i \left( \mathbf{J}^T_a \mathbf{J}_b \right)_{ij} \left( \mathbf{b} - \mathbf{r}_b \right)_j
\end{split}
\end{equation}
Here $\mathbf{a}$ and $\mathbf{b}$ are two inputs, $i$ and
$j$ index their respective features, and $\mathbf{r_a}$ and
$\mathbf{r_b}$ are \textit{reference inputs} which are required to be semantically neutral (i.e. yield a similarity score of zero).
In analogy to \citeauthor{intgrads}, we defined the \textit{integrated Jacobian} $\mathbf{J}_a$ as:
\begin{equation} \label{eq:int_jacobians}
\begin{split}
    (\mathbf{J}_a)_{ki} &= \int_{\alpha=0}^1 \, \frac{\partial 
    \mathbf{e}_k(\mathbf{x}(\alpha))}{\partial \mathbf{x}_i} \, d\alpha \\
    & \approx \frac{1}{N} \, \sum_{n=1}^N \, \frac{\partial \mathbf{e}_k(\mathbf{x}(\alpha_n))}{\partial \mathbf{x}_i},
\end{split}
\end{equation}
which we calculate numerically by summing over interpolation steps along the
straight line between $\mathbf{r}_a$ and
$\mathbf{a}$ given by $\mathbf{x}(\alpha) = \mathbf{r}_a + \alpha (\mathbf{a} - \mathbf{r}_a)$.

\begin{figure}[tb]
    \centering
    \begin{subfigure}{0.99\linewidth}
        \includegraphics[width=\linewidth]{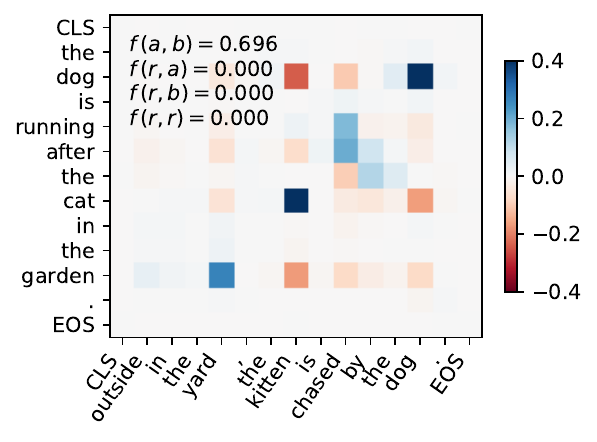}
    \end{subfigure}
    \begin{subfigure}{0.99\linewidth}
        \includegraphics[width=\linewidth]{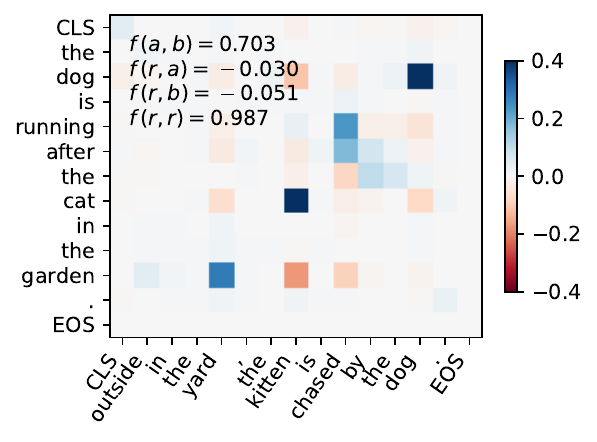}
    \end{subfigure}
    \caption{Attributions for the same example in the \textit{Exact} (top) and \textit{Tuned} (bottom) models. Plots include individual terms from the LHSs of Equation \ref{eq:two_inputs}.}
    \label{fig:attr_agreement}
\end{figure}

The expression inside the sum of the last line in
Equation \ref{eq:two_inputs} is a matrix of all possible feature pairs $ij$
in the two inputs, which we will refer to as $\mathbf{A}$.
It can be reduced to a token--token matrix, as illustrated in Figure\ \ref{fig:attr_agreement}. Provided that the reference inputs $\mathbf{r}$ are dissimilar to any other input sentence $\cdot$ (i.e., $f(\mathbf{r}, \cdot) \!=\! 0$), the last three terms on the left-hand side in Equation \ref{eq:two_inputs} vanish and the sum over the attribution matrix, $\mathbf{A}$,
is exactly equal to the model prediction, $f(\mathbf{a}, \mathbf{b})$.
This is why these attributions can be considered provably correct and we can say they \textit{faithfully} explain which
parts of the inputs the model attends to for a given prediction.

To guarantee the side condition of $f(\mathbf{r}, \cdot) \!=\! 0$, we proposed in \citet{emnlp} to adjust the model
architecture in two ways.  First, we shift all embeddings by the
references, so that $\mathbf{e}(\cdot) = \mathbf{e}'(\cdot) - \mathbf{e}'(\mathbf{r})$,
where $\mathbf{e}'$ is the original encoder and $\,\cdot\,$ is an arbitrary
input. This shift results in references to be mapped onto
the zero vector in the embedding space, which is why
all terms involving $\textbf{r}$ vanish in
Equation \ref{eq:two_inputs}.  Unfortunately, Siamese sentence encoders
typically use cosine as a similarity measure, which
normalizes embedding vectors to unit length. For the zero vector,
normalization is undefined.
This is why, second, we replace the cosine by a dot product in the previous publication.

The application of these adjustments to a model requires fine-tuning. Thus, attributions cannot be derived for the original
model, but only an adapted version of it.  The adjustments also result in a slight decrease of predictive performance (cf. row \textit{Orig.} in Table\ \ref{tab:performance}).  Finally, a dot-product as similarity measure does not guarantee the similarity of a sentence to itself to be one (i.e. maximal). 

\subsection{Proposed Extensions}

In this work we address these two limitations.

\paragraph{Utilizing cosine similarity.}
In Equation\ \ref{eq:int_jacobians}, the integrated Jacobian $\mathbf{J}_a$
results from computing forward- and backward-passes of all
interpolation steps $\mathbf{x}(\alpha_n)$ along the integration
path. However, due to the numerical calculation of the integral, the
closest input to the reference $\mathbf{r}_a$ that is actually ever used
is $\mathbf{x}(\alpha)$ with $\alpha \!=\! \|\mathbf{a}-\mathbf{r}_a\| / N$, 
the first interpolation step for input $\mathbf{a}$. For a
large number of steps $N$, this input may come arbitrarily close to
$\mathbf{r}_a$, but never reaches it. Therefore, in practice we actually never need to normalize the zero embedding-vector $\mathbf{e}(\mathbf{r}_a)$, which
$\mathbf{r}_a$ is mapped to, and we can safely use cosine as a similarity measure.

\paragraph{Approximate References.} 
We can loosen the requirement for references to yield
exact zero similarities, which allows us to avoid the embedding shift. We still use
sequences of padding tokens with the same length as the respective input as references, but we now subtract their emebddings from input embeddings. 
Padding sequences are nevertheless uninformative and should yield similarities close to zero for most input sentences.

As a result, the last three terms on the left-hand side of
Equation \ref{eq:two_inputs} do not vanish any more.  The two \textit{reference similarity} terms involving either input will become close to zero:
$f(\mathbf{a}, \mathbf{r}_b) \approx 0$ and
$f(\mathbf{b}, \mathbf{r}_a) \approx 0$.  The \textit{reference term} will not, but it will become close to one as references should be similar to another, $f(\mathbf{r}_a, \mathbf{r}_b) \approx 1$.  It may not be exactly one, because if the two inputs are of different lengths, so are the two references, and their sentence representations will not be
mapped onto the exact same embedding.


\paragraph{Approximate Attributions.} 
Combining the approximations from above, we obtain the following \textit{approximate
  attribution} method:
\begin{equation} \label{eq:approx_attr}
    f(\mathbf{a}, \mathbf{b}) + 1 \approx \sum_{ij} \left(\mathbf{a} - \mathbf{r}_a \right)_i \left( \mathbf{J}^T_a \mathbf{J}_b \right)_{ij} \left( \mathbf{b} - \mathbf{r}_b \right)_j
\end{equation}

The attribution matrix on the right-hand side no longer exactly
corresponds to the model prediction, because it is now influenced by
the reference term and non-zero reference similarities. A priori, we
cannot tell how both contributions distribute among individual feature
pairs $ij$, and whether they influence the relative order of
attributions.  That being said, the ability to utilize cosine
similarity and the lack of need for an embedding shift obviates the need for fine-tuning to adjust the model architecture, and
Equation \ref{eq:approx_attr} offers a means to compute approximate attributions for off-the-shelf models.

\section{Experiments 1: Analysis of Attributions}
Our approximate attributions do not provide a theoretical guarantee to be correct.
Therefore, in this section, after evaluating predictive performances in different settings, we first quantify the influence of reference contributions to approximate attributions, and then evaluate how well exact and approximate attributions agree.
We work with attributions to layer nine, because they are expressive, while still being accurate with reasonable computational cost \cite{emnlp}.
  
\subsection{Experimental Setup}
We experiment with Siamese sentence transformers trained to predict
semantic textual similarity, and base our evaluation on the
well-established STS benchmark \cite{sts}, consisting of 5749 training, 1500 development and 1379 test sentence pairs from various SemEval%
\footnote{\url{https://semeval.github.io}}
tasks. Our
implementation builds on the \textit{sentence-transformers}%
\footnote{\url{https://www.sbert.net}}
package \citep{sbert}. 
Training details are provided in Appendix \ref{apdx:training}.

\subsection{Predictive Performance}

We first evaluate the performance of Siamese models on the STS data corresponding to different possible configurations for exact and approximate attributions. The aspects discussed in Section\ \ref{sec:exact_attr} give rise to four such configurations,
shown in Table~\ref{tab:performance}: they differ in whether we apply
an embedding shift (Shift), and whether we train the model on the STS
train set (Train).  \textit{Shelf} refers to the unmodified
off-the-shelf version.  \textit{Tuned} undergoes the same training as
the other fine-tuned models but keeps its unmodified architecture. The
\textit{Exact} model introduces the embedding shift enabling exact
attributions. Finally, \textit{Orig.} is the configuration from
\citet{emnlp} with a dot product as the similarity measure.
We evaluate all models\footnote{All models are based on the
  \textit{all-mpnet-base-v2} sentence transformer from \url{https://www.sbert.net/docs/pretrained_models.html}} on the STS
test set using the standard metric of Spearman correlation
between the cosine similarity of embeddings and annotations.

The \textit{Tuned} model achieves the best performance. The
\textit{Orig.} and \textit{Exact} models sacrifice $1.8$ and $0.3$
points in average correlation, respectively. Using the  framework for assessing statistical significance
introduced by \citet{dror}, the superiority of the \textit{Tuned} 
model over the \textit{Exact} one is, however, not significant ($p\!<\!0.05$, details in Appendix \ref{apdx:significance}).
This shows that the embedding shift only minimally harms
the performance when compared against the unmodified model undergoing identical training (\textit{Tuned}).

\begin{table}[tb!]
    \centering
    \begin{tabular}{r c c c c c c}
        \toprule
        \textbf{Model} & \textbf{Shift} & \textbf{Train} & \textbf{Attr.} & $\mathbf{r_S} \!\tiny \times 100$ \\
        \midrule
        \textit{Shelf} & \xmark & \xmark & appr. & 83.4 \\
        \textit{Tuned} & \xmark & \cmark & appr. & \textbf{87.8} \\
        \textit{Exact} & \cmark & \cmark & exact & \underline{87.5} \\
        \textit{Orig.} & \cmark & \cmark & exact & 86.0 \\
        \bottomrule
    \end{tabular}
    \caption{Spearman correlation of cosine similarities between embeddings and target labels in different model settings as described in the text. Top and second best performances are bold / underlined.}
    \label{tab:performance}
\end{table}
\begin{figure}[tb!]
    \centering
    \includegraphics[width=\linewidth]{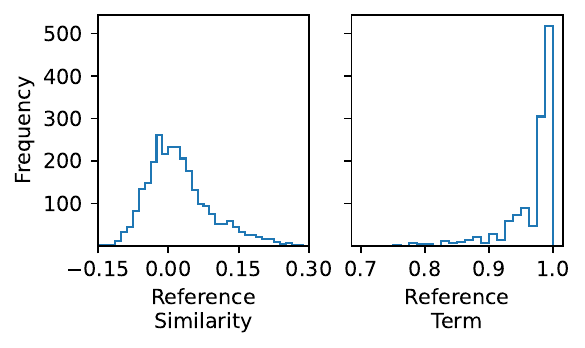}
    \caption{
    Contributions of \textit{reference similarities}, $f(\mathbf{a},\mathbf{r}_b)$ and $f(\mathbf{b},\mathbf{r}_a)$ (left), and the \textit{reference term}, $f(\mathbf{r}_a,\mathbf{r}_b)$ (right), to attributions.}
    \label{fig:ref_sim}
\end{figure}

\subsection{Reference Contributions}
For off-the-shelf models that have not been adapted for the
similarities to the references $f(\mathbf{a},\mathbf{r}_b)$ and $f(\mathbf{b},\mathbf{r}_a)$ to vanish, we can
test how close similarities of inputs to the references actually
are. Figure\ \ref{fig:ref_sim} (left) shows the distribution of
similarities between all STS test set sentences and corresponding reference inputs consisting
of padding tokens.  $85.8 \%$ of all similarities are within an
interval of $\pm 0.1$ around zero.  Thus, in many cases the assumption
for \textit{reference similarities} to be negligible,
$f(\mathbf{r}, \cdot) \approx 0$, may be assumed. However, the width of this distribution also shows that in a substantial fraction of test examples reference similarities are not sufficiently small. Whenever they become non-negligible, they can confound attributions and the approximation of Equation \ref{eq:approx_attr} cannot be assumed safely.
Fortunately, we can perfectly quantify this error case by case by
explicitly computing the reference similarities of both inputs.\\
Similarly, we can evaluate how large the contribution of the \textit{reference term}, $f(\mathbf{r}_a, \mathbf{r}_b)$, to the attributions is.
Figure\ \ref{fig:ref_sim} (right) shows a histogram of all values for
this term.  As expected, they are mostly close
to one.  Only $6.7\%$ of all contributions are smaller than $0.9$.
Different from the \textit{reference similarities} for the two inputs, the
\textit{reference term} is never negligible.

\subsection{Agreement between Exact and Approximate Attributions}
Due to the non-zero reference contributions $f(\mathbf{a},\mathbf{r}_b)$, $f(\mathbf{b},\mathbf{r}_a)$ and $f(\mathbf{r}_a,\mathbf{r}_b)$, the attribution matrix $\mathbf{A}$ can no longer be assumed to exactly reformulate the model prediction $f(\mathbf{a}, \mathbf{b})$, because we cannot tell how the former terms distribute among $\mathbf{A}$ (cf. Equation \ref{eq:two_inputs}). 
In order to evaluate how much reference similarities and the reference
term confound attributions, we compare approximate attributions from
the \textit{Tuned} model against exact ones from the \textit{Exact}
one. For this evaluation, it is important that both models undergo an identical training, with the only difference being that embeddings in the
\textit{Exact} model are shifted. Therefore, we do not compare attributions of the \textit{Shelf} or \textit{Orig.} model in this experiment.

\begin{figure}[tb]
    \centering
    \includegraphics[width=\linewidth]{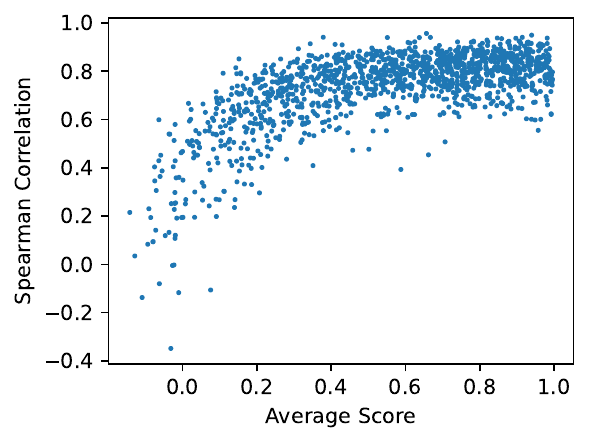}
    \caption{Spearman correlation between attributions from the \textit{Tuned} and \textit{Exact} model for all STS test set pairs (y axis) plotted against the mean predicted similarity of both models (x axis).}
    \label{fig:attr_corr}
\end{figure}

The plots in Figure \ref{fig:attr_agreement} show example attributions of both
models for a random sentence pair from the
STS test set. As expected, in the \textit{Exact} model both reference
similarities and the reference term vanish, while in the
\textit{Tuned} one, the former come close to zero and the latter is
approximately one.  Some attributions are quite different, however, a
general pattern appears to be rather well preserved.  We evaluate how
consistently this is the case by computing attributions from both
models for all sentences in the STS test set and compare them.  We are
also interested how the agreement of attributions behaves as a
function of similarity score.  In Figure\ \ref{fig:attr_corr}, we plot
Spearman correlation values of attributions to layer eleven against
the average similarity score predicted by the two models.  The
correlation steadily increases with higher similarity scores. For
scores $s\!>\!0.75$ it reaches $r_S=0.81\!\pm\!0.07$.

\begin{figure}[tb]
    \centering
    \includegraphics[width=\linewidth]{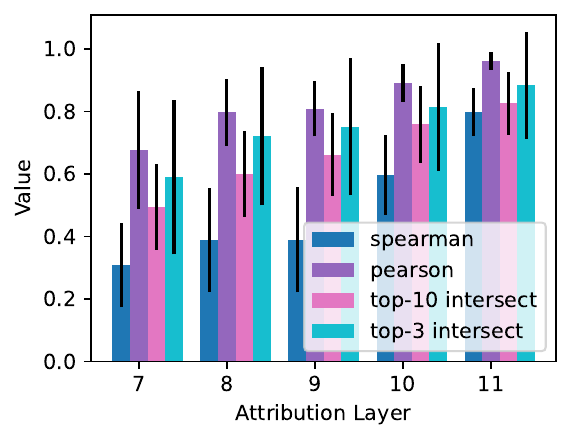}
    \caption{Agreement between attributions by the \textit{Tuned} and \textit{Exact} model. We compute Spearman and Pearson correlations, as well as the intersections between the top-3 and top-10 attributions for different layers and similarity scores $s>0.5$.}
    \label{fig:layer_attr_corr}
\end{figure}

We repeat this experiment for attributions to all layers down to the
seventh, for which we have previously found attributions to be sufficiently
accurate with $N\!<\!200$ \cite{emnlp}.  Figure\ \ref{fig:layer_attr_corr} summarizes the results for similarity scores $s>0.5$. Spearman correlation declines to $r_S=0.60\pm0.13$ and $r_S=0.40\pm0.17$ in layer ten and nine,
respectively.
We note that Spearman correlation only regards the rank
of attributions and will be strongly influenced by small attributions,
which may be dominated by noise and do not interest us very much.
Pearson correlation, on the other hand, which remains relatively high
with $r_P=0.80\pm0.11$ in layer eight, is technically not suitable because
we cannot presuppose a linear relation between attributions.
We are mostly interested in the agreement of attributions that stand out.  
Therefore, we also evaluate the overlap among the top ten (and three, shown in parentheses) attributions in all pairs.  
The Jaccard coefficient starts at $0.83\!\pm\!0.10$ ($0.88\!\pm\!0.17$) in layer eleven and
decreases to $0.60\!\pm\!0.14$ ($0.75\!\pm\!0.20$) in layer eight.

These results show that approximate attributions
are trustworthy for very deep layers.  Attributions to deeper
intermediate representations may still provide interesting insights,
but must be interpreted with caution and cannot be taken to be
completely reliable. The results also show that care must be applied regarding dissimilar sentence pairs, because for very low scores, approximate attributions do not agree with exact ones.

\subsection{Positive and negative attributions}\label{ssec:pos-neg}
Intuitively, pairs of tokens with congruent semantics, which make a
pair of sentences more semantically similar, should positively
contribute to the similarity score and receive positive attribution
scores. Conversely, pairs of tokens that contradict each other should
be assigned negative attributions in order to push the similarity
score towards zero, cf.\ the effect of \textit{not} in Figure\
\ref{fig:negation_expl}.
Examination of attribution matrices shows, however, that this
 scenario is quite rare and we often fail to see noticeable
negative attributions where we expect them.

A possible reasons for this behavior is that models  tend to \enquote{overshoot} 
the contributions of semantically congruent tokens and
need to balance them out by assigning negative contributions to neutral token pairs (unlike 
the final scores, token-pair contributions can take any value).


\begin{figure}[tb!]
    \centering
    \includegraphics[width=\linewidth]{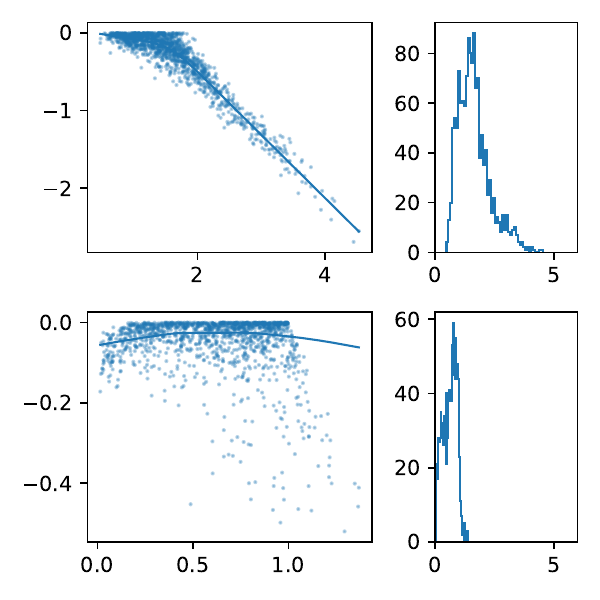}
    \caption{The relationships (with LOWESS smoothing)
    between sums of positive and negative elements of attribution
    matrices computed on the STS test set using the \textit{Shelf} and the \textit{Exact} model (left pane of top and bottom row, respectively) and the distribution of sums of positive elements in these matrices (right pane).}
    \label{fig:pos-neg-sums}
\end{figure}

In order to test if this is the case, we separately extract  the sums of all
positive and all negative elements of the attribution matrices computed based on the sentence
pairs from the STS test set using two similarity models. The relationship between the sums
of positive and negative token-pair attributions across sentences is shown in 
Figure~\ref{fig:pos-neg-sums}.
Both models demonstrate
cases when positive attributions sum to more than the score maximum, which is 1 for the exact
model and 2 for the off-the-shelf model
(cf.\ Equations.~\ref{eq:two_inputs}
and \ref{eq:approx_attr}), thus demanding a proportional total negative
contribution. However, this analysis also shows a difference between the exact model
and the approximate model: we see approximate attributions computed on the basis of the off-the-shelf model summing 
to more than 2 much more frequently than exact attributions summing to more than 1.
We cannot tell whether this effect is an artifact of the approximate attribution
method or whether the model itself actually assigns such large contributions, while the weights of fine-tuned exact model become normalized.
Overall, the data show that, unfortunately, negative attributions are not entirely reliable
even in the exact attribution setting, given that positive attributions sometimes sum
to more than 1, and in the approximate setting the proportion of these cases is higher.

\section{Experiments 2: Analysis of Sentence Transformers}

The attributions derived by our method let us directly analyze the decision making process inside STs for the first time.
In this section, we extend the analysis to
concrete levels of linguistic structures including syntactic functions, negation, adjectives, and general lexical effects. 

\subsection{Syntactic Relations}
\begin{figure}
    \centering
    \includegraphics[width=\linewidth]{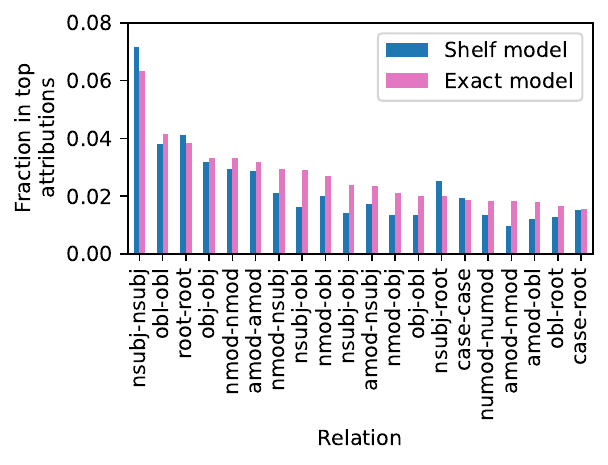}
    \caption{Distribution of relations between syntactic functions of word-pairs in the top 10\% of all attributions for instances in the STS test set.}
    \label{fig:dep_rel}
\end{figure}

\citet{emnlp} evaluated which relations between parts of speech Siamese language models typically consult. We extend this analysis to relations between the syntactic functions of words.
Using a Universal Dependencies parser,\footnote{We use Stanza \citep{qi2020stanza}.} 
we obtain parse trees for all sentence pairs from the STS test set, and replace labels of multi-word expressions and coordinated constructions with the label of their closest parent that is not phrase internal.
On the attribution side, we combine token- to word-attributions by averaging.
We then extract syntactic relations of the top $10\%$ of all attributions in every sentence pair. 
Figure\ \ref{fig:dep_rel} shows a distribution of the most attributed relations in our \textit{Exact} model and the  (off-the-) \textit{Shelf} model.

As one may expect, subject (\texttt{nsubj}), predicate (\texttt{root} marks the predicate of the 
main clause), direct object (\texttt{obj}), and oblique (\texttt{obl}) relations appear among the top attributions. 
Notably, top-contributing pairs are based on words with identical syntactic function, 
which suggests that models begin by matching major syntactic roles before considering mixed relations. 
Same-function word pairs also show high agreement between models.
The two models do not agree so well on attributions to word-pairs of different function.
The \textit{Exact} model tends to attribute to subject--object (\texttt{nsubj}--\texttt{obj}) pairs much 
more often. 
The opposite is true for subject--predicate (\texttt{nsubj}--\texttt{root}) relations, which the \textit{Shelf} model attributes more often than any other mixed relation.
In the exact model, somewhat surprisingly, this relation only appears on rank 14.\\
With a fraction of around $7\%$ subject--subject attributions are by far the most frequent.
Nevertheless, this is not a large share of all top attributions, and the rest of the distribution 
does not decline steeply.
Therefore, we can conclude that the models regard a wide range of relations between syntactic roles 
and do not overly focus on specific ones. At the same time, the relative important of participant-like
elements supports that the conclusions reached by \citet{nikolaev-pado-2023-representation} for synthetic sentences generalize to natural text. 

\subsection{Negation}\label{ssec:negation}
It is a well-known fact that sentence transformers do not handle negation well \cite{negation}.
We use our attribution method to seek a deeper understanding of this phenomenon.
From the STS test set, we extract 87 sentences that contain a simple \textit{not}-negation.
We then derive attributions for the similarity to the identical but non-negated sentence and compute the total attribution to the \textit{not}-token.

\begin{figure}[t]
    \centering
    \includegraphics[width=\linewidth]{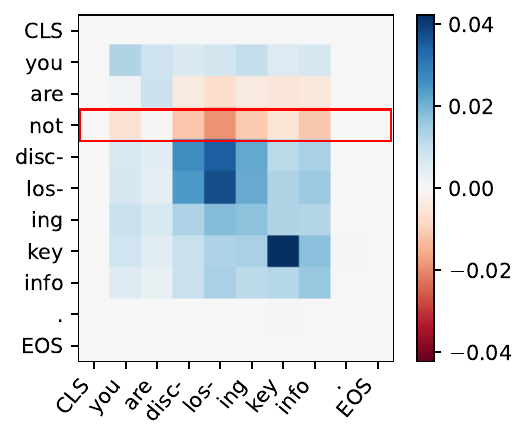}
    \caption{Attribution matrix for a sentence paired with its negation (\textit{Exact} model). The red box marks the  contribution of the negation. This is
    a rare example of a clear negative contribution (cf.\ Figure\ \ref{fig:not_attr_distr}).}
    \label{fig:negation_expl}
\end{figure}

\begin{figure}[t!]
    \centering
    \includegraphics[width=\linewidth]{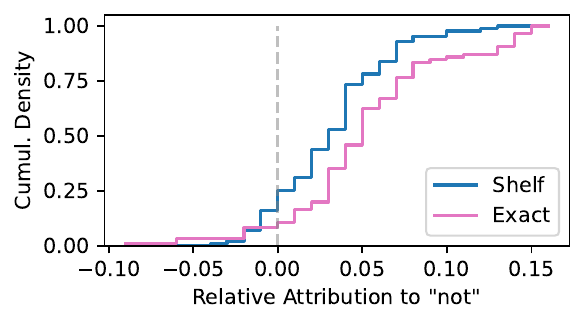}
    \caption{Cumulative distributions of total attributions to the \textit{not}-token when computing the similarity of a negated sentence to its non-negated version.}
    \label{fig:not_attr_distr}
\end{figure}

The negation should show a negative contribution in the attribution; Figure\ \ref{fig:negation_expl} shows an example where this is actually the case.
However, the distribution of attributions in Figure\ \ref{fig:not_attr_distr} shows that this is not the usual behaviour.
In the \textit{Shelf} (\textit{Exact}) model
only approximately $16\%$ ($9\%$) of all \textit{not}-attributions are negative. 
In $90\%$ of the cases relative attributions to the \textit{not}-token account for less than $8\%$ ($14\%$) of the prediction.
This provides additional evidence for the fact that sentence transformers mostly ignore negation.

\subsection{Adjectives as Predicates}
As another test of the STs' ability to model polarity,
we construct adjective triplets. These combine an anchor adjective with one synonymous and one opposite adjective, e.g. \textit{pretty} with \textit{beautiful} and \textit{ugly}.
From a total of 23 such triplets, we then build a synthetic data set consisting of two sentence pairs per triplet (Appendix \ref{apdx:adj}) built from the same sentence template. The sentences  differ only in 
the adjective position: One sentence combines the original
and the synonymous adjective (\textit{This house is beautiful.}, \textit{This house is pretty.}), one the original with the opposite one (\textit{This house is pretty.}, \textit{This house is ugly.}).

\begin{figure}[t]
    \centering
    \begin{subfigure}{.99\linewidth}
        \centering
        \includegraphics[width=\linewidth]{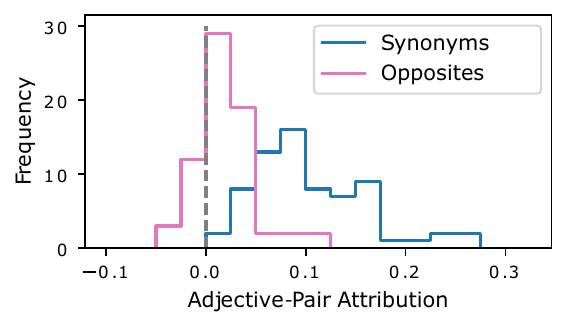}
    \end{subfigure}
    \begin{subfigure}{.99\linewidth}
        \centering 
        \includegraphics[width=\linewidth]{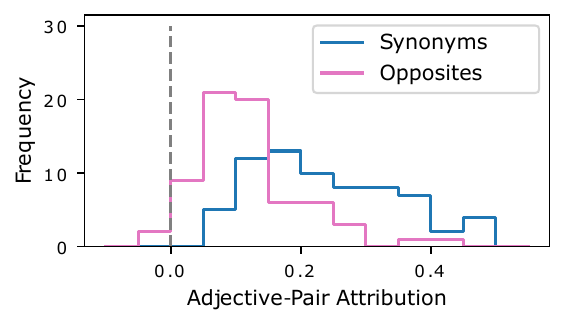}
    \end{subfigure}
    \caption{Histograms of attributions to synonymous and opposite adjective pairs from the \textit{Exact} (top) and the \textit{Shelf} model (bottom).}
    \label{fig:adj_attr_distr}
\end{figure}

We then compute attribution matrices for the two sentence-pairs from every instance, combine token-level to word-level attributions by averaging and evaluate the attributions to the respective adjective pairs. We expect the synonymous pairs to contribute pronounced positive attributions to sentence-similarities. Opposite pairs, on the other hand, result in two sentences with opposing meaning. One may expect that respective adjective-pairs should, hence, receive negative attributions.
However, we find that this is not typically the case.
In Figure\ \ref{fig:adj_attr_distr}, we plot histograms of the attributions to synonymous and opposite adjective pairs for both the \textit{Exact} and the \textit{Shelf} model.

In both cases the distributions show that opposite adjective pairs, generally, do receive lower, but only rarely negative attributions.

\subsection{Lexical effects}
\label{ssec:lexical-effects}


Finally, we investigate whether attributions are lexically
biased, i.e.\ whether similarity scores produced by SEs are sensitive to the exact
lexical choice. E.g., given a pair of sentences like \textit{A puppy was born in X.}
vs.\ \textit{How many hurricanes occur in X each year?}, intuitively we do not
expect the similarity score to noticeably vary with the choice of X. 
However, the \textit{Shelf} model predicts scores above 0.3 when 
X is in {\textit{Auckland}, \textit{Cambodia}, \textit{Granville}} but only 0.13 for \textit{the USA} and 0.19 for \textit{Europe}.

In order to study this more systematically, we use the QQP
dataset\footnote{\url{https://quoradata.quora.com/First-Quora-Dataset-Release-Question-Pairs}}
containing more than 400k question pairs\footnote{See Appendix~\ref{app:lexical-effects} for experimental details.}
and record values of all matrix cells corresponding to same-token
pairs. We then extract all attributions for words appearing 30 and
more times and assign them ranks based on their average attributions.

\begin{figure}[t]
    \centering
    \includegraphics[width=1\linewidth]{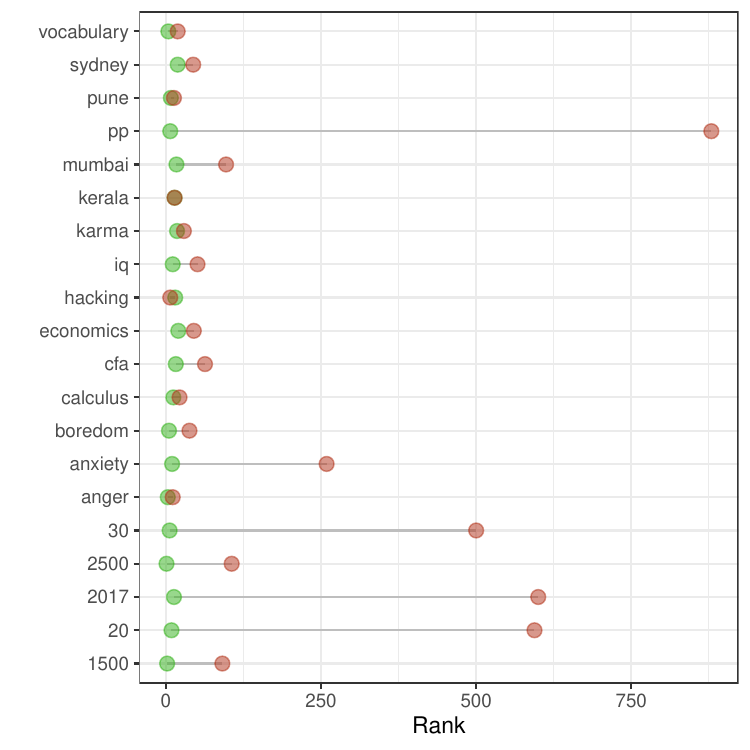}
    \caption{Top-ranking words by their same-token attributions over the QQP dataset sentence pairs. 
    X-axis: ranks (\textit{Exact} model ranks shown in green, \textit{Shelf}
    model ranks in brown).}
    \label{fig:lexical-ranks-comparison}
\end{figure}

As should be expected, both the \textit{Exact}
model and the \textit{Shelf} model pay little attention to EOS, CLS, and 
punctuation signs, which obtain the lowest ranks in both models.
As for the top ranks, both models give high ranks to certain place names
(\textit{Kerala}, \textit{Pune}), words describing emotions (\textit{anger},
\textit{boredom}), and a seemingly random assortment of other words corresponding
to different question topics (\textit{hacking}, \textit{vocabulary}, 
\textit{furniture}). Interestingly, while the \textit{Exact} model also assigns very
high importance to particular numbers (\textit{2500}, \textit{1500}, etc.),
the \textit{Shelf} is less sensitive to them (the top number token, \textit{1500},
has rank 91). Comparison of ranks for top tokens is shown in 
Figure~\ref{fig:lexical-ranks-comparison}. Overall, the attribution ranks show high 
agreement (Spearman's \(r \!=\! 0.81\)) between the two models, and the standard deviations
for the contributions are rather low (cf.\ 
Table~\ref{tab:lexical-attributions-exact} and \ref{tab:lexical-attributions-shelf} 
and Figure~\ref{fig:top-30-lexical-densities} in the appendix), which shows that 
lexical effects are both strong and consistent.

\section{Conclusion} \label{sec:conclusion}
The updates to our original method proposed in this paper (i)~result in a Siamese Transformer with exact attribution ability to retain the predictive performance of the equivalent unmodified model, and (ii)~enable a way to compute approximate attributions for Siamese encoders which can be directly applied to off-the-shelf models without the need for fine-tuning.
Unlike their exact counterparts, these approximate attributions do not come with the theoretical guarantee to exactly reflect the model prediction.  Our evaluation, however, shows that for deep intermediate representations they are reliable to a certain extent and often agree with exact attributions.

Analyses carried out based on our attributions show that Siamese transformers primarily match subjects, predicates and objects but also considering different syntactic relations. 
They mostly do not attend to negation and often assign small yet positive contributions to semantic opposites. On a lexical level, some words always obtain high attributions with small variance whenever they appear.

On the methdological level, we suggest that due to the practicality of approximate attributions, they may be used to obtain a first round of insights into off-the-shelf models.
Whenever reliable attributions of predictions are required, however, an exact attribution model should be employed.
Therefore, an interesting future perspective will be to train large Siamese models with exact attribution ability from scratch.

\section{Limitations}

We first emphasize that in this paper a central limitation of our original attribution method for Siamese encoders \cite{emnlp}, namely that a dot-product instead of cosine needs to be used as a similarity measure, is removed.
This results in the fact that self-similarity of sentences is guaranteed to be one, instead of being unbound.

The central limitation of approximate attributions for off-the-shelf Siamese Encoders in this paper is that they do not exactly reflect model predictions, which is elaborately discussed above.

A second important limitation remain the high computational costs for attributions to input and shallow intermediate representations. With our available computational resources and the current implementation accurate attributions to shallow layers are not tractable \cite{emnlp}.
In the future it will also be important to look into potential options to increase the efficiency for the computation of these attributions. 

Finally, deeper intermediate representations in transformer models are contextualized and hence do not represent the associated token alone, but its context.
In the future it will also be interesting to investigate the relation between attributions to different layers and contextualization. 
\FloatBarrier

\bibliography{anthology,custom}

\begin{thebibliography}{24}
\expandafter\ifx\csname natexlab\endcsname\relax\def\natexlab#1{#1}\fi

\bibitem[{Atanasova et~al.(2020)Atanasova, Simonsen, Lioma, and
  Augenstein}]{atanasova}
Pepa Atanasova, Jakob~Grue Simonsen, Christina Lioma, and Isabelle Augenstein.
  2020.
\newblock \href {https://doi.org/10.18653/v1/2020.emnlp-main.263} {A diagnostic
  study of explainability techniques for text classification}.
\newblock In \emph{Proceedings of the 2020 Conference on Empirical Methods in
  Natural Language Processing (EMNLP)}, pages 3256--3274, Online. Association
  for Computational Linguistics.

\bibitem[{Bastings and Filippova(2020)}]{bastings}
Jasmijn Bastings and Katja Filippova. 2020.
\newblock \href {https://doi.org/10.18653/v1/2020.blackboxnlp-1.14} {The
  elephant in the interpretability room: Why use attention as explanation when
  we have saliency methods?}
\newblock In \emph{Proceedings of the Third BlackboxNLP Workshop on Analyzing
  and Interpreting Neural Networks for NLP}, pages 149--155, Online.
  Association for Computational Linguistics.

\bibitem[{Cer et~al.(2017)Cer, Diab, Agirre, Lopez-Gazpio, and Specia}]{sts}
Daniel Cer, Mona Diab, Eneko Agirre, I{\~n}igo Lopez-Gazpio, and Lucia Specia.
  2017.
\newblock \href {https://doi.org/10.18653/v1/S17-2001} {{S}em{E}val-2017 task
  1: Semantic textual similarity multilingual and crosslingual focused
  evaluation}.
\newblock In \emph{Proceedings of the 11th International Workshop on Semantic
  Evaluation ({S}em{E}val-2017)}, pages 1--14, Vancouver, Canada. Association
  for Computational Linguistics.

\bibitem[{Clark et~al.(2019)Clark, Khandelwal, Levy, and Manning}]{clark}
Kevin Clark, Urvashi Khandelwal, Omer Levy, and Christopher~D. Manning. 2019.
\newblock \href {https://doi.org/10.18653/v1/W19-4828} {What does {BERT} look
  at? an analysis of {BERT}{'}s attention}.
\newblock In \emph{Proceedings of the 2019 ACL Workshop BlackboxNLP: Analyzing
  and Interpreting Neural Networks for NLP}, pages 276--286, Florence, Italy.
  Association for Computational Linguistics.

\bibitem[{Conneau et~al.(2017)Conneau, Kiela, Schwenk, Barrault, and
  Bordes}]{conneau}
Alexis Conneau, Douwe Kiela, Holger Schwenk, Lo{\"\i}c Barrault, and Antoine
  Bordes. 2017.
\newblock \href {https://doi.org/10.18653/v1/D17-1070} {Supervised learning of
  universal sentence representations from natural language inference data}.
\newblock In \emph{Proceedings of the 2017 Conference on Empirical Methods in
  Natural Language Processing}, pages 670--680, Copenhagen, Denmark.
  Association for Computational Linguistics.

\bibitem[{Danilevsky et~al.(2020)Danilevsky, Qian, Aharonov, Katsis, Kawas, and
  Sen}]{xnlp}
Marina Danilevsky, Kun Qian, Ranit Aharonov, Yannis Katsis, Ban Kawas, and
  Prithviraj Sen. 2020.
\newblock \href {https://aclanthology.org/2020.aacl-main.46} {A survey of the
  state of explainable {AI} for natural language processing}.
\newblock In \emph{Proceedings of the 1st Conference of the Asia-Pacific
  Chapter of the Association for Computational Linguistics and the 10th
  International Joint Conference on Natural Language Processing}, pages
  447--459, Suzhou, China. Association for Computational Linguistics.

\bibitem[{Doshi-Velez and Kim(2017)}]{doshi}
Finale Doshi-Velez and Been Kim. 2017.
\newblock \href {https://arxiv.org/abs/1702.08608} {Towards a rigorous science
  of interpretable machine learning}.
\newblock \emph{arXiv:1702.08608}.

\bibitem[{Dror et~al.(2019)Dror, Shlomov, and Reichart}]{dror}
Rotem Dror, Segev Shlomov, and Roi Reichart. 2019.
\newblock \href {https://doi.org/10.18653/v1/P19-1266} {Deep dominance - how to
  properly compare deep neural models}.
\newblock In \emph{Proceedings of the 57th Annual Meeting of the Association
  for Computational Linguistics}, pages 2773--2785, Florence, Italy.
  Association for Computational Linguistics.

\bibitem[{Kobayashi et~al.(2024)Kobayashi, Kuribayashi, Yokoi, and
  Inui}]{kobayashi2023analyzing}
Goro Kobayashi, Tatsuki Kuribayashi, Sho Yokoi, and Kentaro Inui. 2024.
\newblock \href {https://openreview.net/forum?id=mYWsyTuiRp} {Analyzing
  feed-forward blocks in transformers through the lens of attention map}.
\newblock In \emph{Proceedings of the Twelfth International Conference on
  Learning Representations}, Vienna, Austria.

\bibitem[{Li et~al.(2016)Li, Chen, Hovy, and Jurafsky}]{li}
Jiwei Li, Xinlei Chen, Eduard Hovy, and Dan Jurafsky. 2016.
\newblock \href {https://doi.org/10.18653/v1/N16-1082} {Visualizing and
  understanding neural models in {NLP}}.
\newblock In \emph{Proceedings of the 2016 Conference of the North {A}merican
  Chapter of the Association for Computational Linguistics: Human Language
  Technologies}, pages 681--691, San Diego, California. Association for
  Computational Linguistics.

\bibitem[{MacAvaney et~al.(2022)MacAvaney, Feldman, Goharian, Downey, and
  Cohan}]{macavaney}
Sean MacAvaney, Sergey Feldman, Nazli Goharian, Doug Downey, and Arman Cohan.
  2022.
\newblock \href {https://doi.org/10.1162/tacl_a_00457} {{ABNIRML}: Analyzing
  the behavior of neural {IR} models}.
\newblock \emph{Transactions of the Association for Computational Linguistics},
  10:224--239.

\bibitem[{Murdoch et~al.(2019)Murdoch, Singh, Kumbier, Abbasi-Asl, and
  Yu}]{murdoch}
W~James Murdoch, Chandan Singh, Karl Kumbier, Reza Abbasi-Asl, and Bin Yu.
  2019.
\newblock Definitions, methods, and applications in interpretable machine
  learning.
\newblock \emph{Proceedings of the National Academy of Sciences},
  116(44):22071--22080.

\bibitem[{Möller et~al.(2023)Möller, Nikolaev, and Padó}]{emnlp}
Lucas Möller, Dmitry Nikolaev, and Sebastian Padó. 2023.
\newblock \href {https://doi.org/10.18653/v1/2023.emnlp-main.980} {An
  attribution method for {S}iamese encoders}.
\newblock In \emph{Proceedings of the 2023 Conference for Empirical Methods in
  Natural Language Processing}, pages 15818--15827, Singapore. Association for
  Computational Linguistics.

\bibitem[{Nikolaev and Pad{\'o}(2023)}]{nikolaev-pado-2023-representation}
Dmitry Nikolaev and Sebastian Pad{\'o}. 2023.
\newblock \href {https://aclanthology.org/2023.eacl-main.268} {Representation
  biases in sentence transformers}.
\newblock In \emph{Proceedings of the 17th Conference of the European Chapter
  of the Association for Computational Linguistics}, pages 3701--3716,
  Dubrovnik, Croatia. Association for Computational Linguistics.

\bibitem[{Opitz and Frank(2022)}]{opitz}
Juri Opitz and Anette Frank. 2022.
\newblock \href {https://aclanthology.org/2022.aacl-main.48} {{SBERT} studies
  meaning representations: Decomposing sentence embeddings into explainable
  semantic features}.
\newblock In \emph{Proceedings of the 2nd Conference of the Asia-Pacific
  Chapter of the Association for Computational Linguistics and the 12th
  International Joint Conference on Natural Language Processing (Volume 1: Long
  Papers)}, pages 625--638, Online only. Association for Computational
  Linguistics.

\bibitem[{Qi et~al.(2020)Qi, Zhang, Zhang, Bolton, and Manning}]{qi2020stanza}
Peng Qi, Yuhao Zhang, Yuhui Zhang, Jason Bolton, and Christopher~D. Manning.
  2020.
\newblock \href {https://nlp.stanford.edu/pubs/qi2020stanza.pdf} {Stanza: A
  {Python} natural language processing toolkit for many human languages}.
\newblock In \emph{Proceedings of the 58th Annual Meeting of the Association
  for Computational Linguistics: System Demonstrations}.

\bibitem[{Reimers and Gurevych(2019)}]{sbert}
Nils Reimers and Iryna Gurevych. 2019.
\newblock \href {https://doi.org/10.18653/v1/D19-1410} {Sentence-{BERT}:
  Sentence embeddings using {S}iamese {BERT}-networks}.
\newblock In \emph{Proceedings of the 2019 Conference on Empirical Methods in
  Natural Language Processing and the 9th International Joint Conference on
  Natural Language Processing (EMNLP-IJCNLP)}, pages 3982--3992, Hong Kong,
  China. Association for Computational Linguistics.

\bibitem[{Rogers et~al.(2020)Rogers, Kovaleva, and Rumshisky}]{rogers}
Anna Rogers, Olga Kovaleva, and Anna Rumshisky. 2020.
\newblock \href {https://doi.org/10.1162/tacl_a_00349} {A primer in
  {BERT}ology: What we know about how {BERT} works}.
\newblock \emph{Transactions of the Association for Computational Linguistics},
  8:842--866.

\bibitem[{Rudin(2019)}]{rudin}
Cynthia Rudin. 2019.
\newblock Stop explaining black box machine learning models for high stakes
  decisions and use interpretable models instead.
\newblock \emph{Nature machine intelligence}, 1(5):206--215.

\bibitem[{Sundararajan et~al.(2017)Sundararajan, Taly, and Yan}]{intgrads}
Mukund Sundararajan, Ankur Taly, and Qiqi Yan. 2017.
\newblock Axiomatic attribution for deep networks.
\newblock In \emph{Proceedings of the 34th International Conference on Machine
  Learning - Volume 70}, page 3319–3328. JMLR.org.

\bibitem[{Thakur et~al.(2021)Thakur, Reimers, R{\"u}ckl{\'e}, Srivastava, and
  Gurevych}]{beir}
Nandan Thakur, Nils Reimers, Andreas R{\"u}ckl{\'e}, Abhishek Srivastava, and
  Iryna Gurevych. 2021.
\newblock \href {https://openreview.net/forum?id=wCu6T5xFjeJ} {{BEIR}: A
  heterogeneous benchmark for zero-shot evaluation of information retrieval
  models}.
\newblock In \emph{Thirty-fifth Conference on Neural Information Processing
  Systems Datasets and Benchmarks Track (Round 2)}.

\bibitem[{Vahtola et~al.(2022)Vahtola, Creutz, and Tiedemann}]{negation}
Teemu Vahtola, Mathias Creutz, and J{\"o}rg Tiedemann. 2022.
\newblock \href {https://doi.org/10.18653/v1/2022.blackboxnlp-1.20} {It is not
  easy to detect paraphrases: Analysing semantic similarity with antonyms and
  negation using the new {S}em{A}nto{N}eg benchmark}.
\newblock In \emph{Proceedings of the Fifth BlackboxNLP Workshop on Analyzing
  and Interpreting Neural Networks for NLP}, pages 249--262, Abu Dhabi, United
  Arab Emirates (Hybrid). Association for Computational Linguistics.

\bibitem[{Voita et~al.(2019)Voita, Talbot, Moiseev, Sennrich, and
  Titov}]{voita}
Elena Voita, David Talbot, Fedor Moiseev, Rico Sennrich, and Ivan Titov. 2019.
\newblock \href {https://doi.org/10.18653/v1/P19-1580} {Analyzing multi-head
  self-attention: Specialized heads do the heavy lifting, the rest can be
  pruned}.
\newblock In \emph{Proceedings of the 57th Annual Meeting of the Association
  for Computational Linguistics}, pages 5797--5808, Florence, Italy.
  Association for Computational Linguistics.

\bibitem[{Wiegreffe and Pinter(2019)}]{wiegreffe}
Sarah Wiegreffe and Yuval Pinter. 2019.
\newblock \href {https://doi.org/10.18653/v1/D19-1002} {Attention is not not
  explanation}.
\newblock In \emph{Proceedings of the 2019 Conference on Empirical Methods in
  Natural Language Processing and the 9th International Joint Conference on
  Natural Language Processing (EMNLP-IJCNLP)}, pages 11--20, Hong Kong, China.
  Association for Computational Linguistics.

\end{thebibliography}

\FloatBarrier

\appendix

\section{Training Details} \label{apdx:training}
We fine-tune all models in the same way and mostly stick to the default setting that is used in the \textit{sentence-transformers} package. The batch size is $16$, and wen run all trainings for five epochs. We use an AdamW-optimizer with a weight decay of $0.1$ and learning rate of $2\times10^{-5}$, taking $10\%$ of the data for linear warm-up.

\section{Significance Testing} \label{apdx:significance}
\citet{dror} introduced a framework that is particularly suitable to test the significance of 
performance improvements between deep learning models.
We apply this test on the distribution of squared errors between predictions and targets on the STS test set 
(MSE is used as a loss function at training time).
We set the tests $\epsilon$-parameter to the suggested value of $\epsilon\!=\!0.3$ and choose a 
significance level of $p\!<\!0.05$, which is not an overly strict criterion for superiority.

\section{Adjective Sentences} \label{apdx:adj}

\begin{figure}
    \centering
    \includegraphics[width=\linewidth]{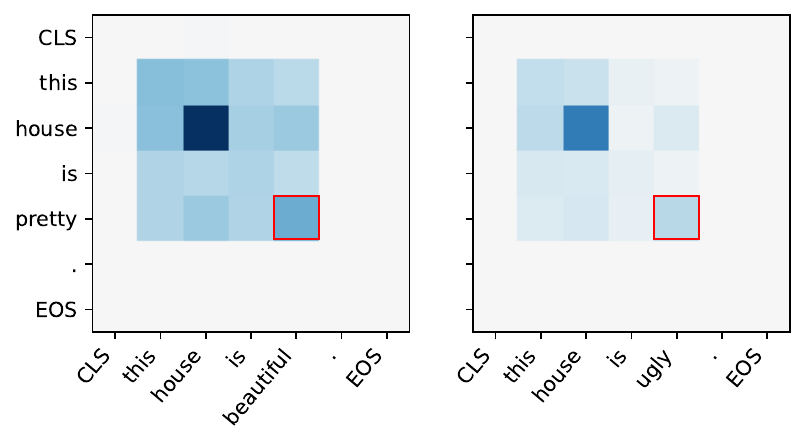}
    \caption{Example attributions for adjective sentences.}
    \label{fig:adj_attr_expl}
\end{figure}

Table \ref{tab:adjectives} lists the 23 adjective triplets that we use to construct sentence pairs.
\begin{table}[]
    \centering
    \begin{tabular}{lll}
        \toprule
         Anchor & Synonym & Opposite  \\
         \midrule
         beautiful & pretty & ugly \\
         ugly & hideous & beautiful \\
         small & little & big \\
         big & huge & small \\
         gigantic & enormous & tiny \\
         tiny & minuscule & enormous \\
         old & elderly & young \\
         young & youthful & old \\
         difficult & hard & easy \\
         simple & easy & difficult \\
         thorough & comprehensive & erroneous \\
         faulty & erroneous & thorough \\
         dirty & messy & clean \\
         clean & tidy & dirty \\
         heavy & massive & light \\
         common & normal & unusual \\
         untypical & unusual & normal \\
         boring & dull & interesting \\
         exciting & interesting & boring \\
         calm & peaceful & hectic \\
         chaotic & hectic & calm \\
         balanced & equal & uneven \\
         unequal & uneven & balanced\\
         \bottomrule
    \end{tabular}
    \caption{Adjective triplets used for our synthetic dataset}
    \label{tab:adjectives}
\end{table}
From these triplets we construct sentence tuples like the following: (\textit{This house is beautiful.}, \textit{This house is pretty.}) and (\textit{This house is beautiful.}, \textit{This house is ugly.}).
Figure \ref{fig:adj_attr_expl} shows attribution matrices for this example and marks the adjective attributions that we compare in red.

\section{Lexical Effects}
\label{app:lexical-effects}

\begin{figure}[t]
    \centering
    \includegraphics[width=1\linewidth]{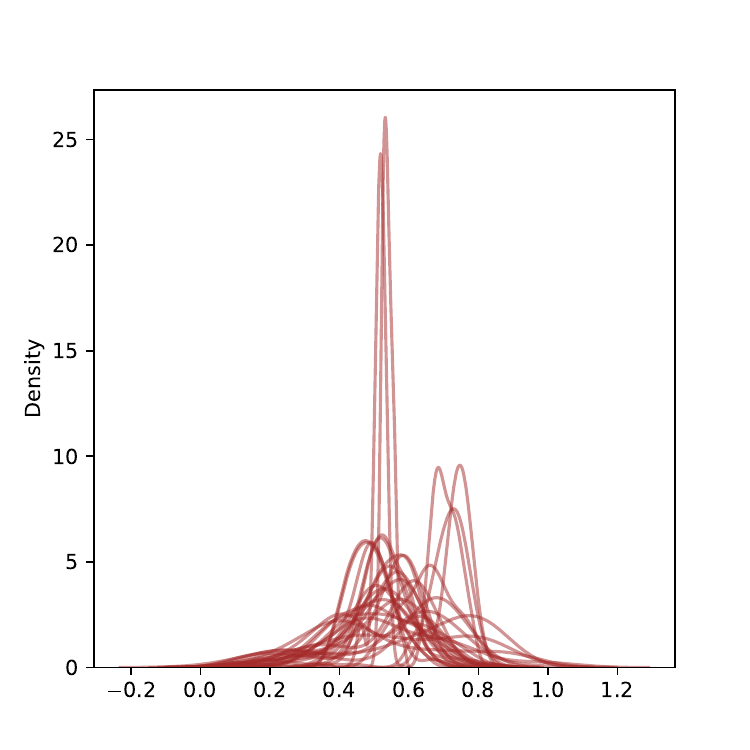}
    \caption{Densities of same-token-pair contributions of 30 lexical items with the
    highest average contribution.}
    \label{fig:top-30-lexical-densities}
\end{figure}

We compute attribution
matrices for 148315 sentence pairs at level 8; \textit{N} = 100. Due to
time constraints we could not compute attributions for all sentence pairs for both
models. However, we computed them for the \textit{Shelf} model, and the results
are nearly identical to those achieved on the subsample, with Spearman's 
\(r > 0.99\). Top-20 and bottom-20 tokens by average contribution to the similarity score in 
identical pairs for the two models are shown in Table~\ref{tab:lexical-attributions-exact} and
\ref{tab:lexical-attributions-shelf}. Densities of same-token-pair contributions of 30 lexical items 
with the highest average contribution are shown in Figure~\ref{fig:top-30-lexical-densities}.

\begin{table}[t]
\centering
\begin{tabular}{@{}lll@{}}
\toprule
\textbf{Word} & \textbf{Mean} & \textbf{StDev} \\ \midrule
2500 & 0.399 & 0.169 \\
1500 & 0.370 & 0.116 \\
anger & 0.236 & 0.096 \\
vocabulary & 0.218 & 0.087 \\
boredom & 0.216 & 0.051 \\
30 & 0.212 & 0.178 \\
pp & 0.212 & 0.185 \\
pune & 0.205 & 0.110 \\
20 & 0.203 & 0.142 \\
anxiety & 0.199 & 0.116 \\
iq & 0.191 & 0.122 \\
calculus & 0.190 & 0.176 \\
2017 & 0.189 & 0.090 \\
kerala & 0.182 & 0.067 \\
hacking & 0.181 & 0.105 \\
cfa & 0.178 & 0.120 \\
mumbai & 0.174 & 0.112 \\
karma & 0.171 & 0.086 \\
sydney & 0.170 & 0.100 \\
economics & 0.168 & 0.115 \\ \midrule
very & 0.003 & 0.005 \\
described & 0.003 & 0.003 \\
( & 0.003 & 0.004 \\
 & 0.003 & 0.004 \\
" & 0.003 & 0.004 \\
hear & 0.003 & 0.004 \\
because & 0.003 & 0.003 \\
) & 0.002 & 0.003 \\
, & 0.002 & 0.003 \\
. & 0.002 & 0.005 \\
@ & 0.002 & 0.009 \\
ones & 0.002 & 0.001 \\
{[} & 0.002 & 0.003 \\
\{ & 0.001 & 0.001 \\
{]} & 0.001 & 0.001 \\
\_ & 0.001 & 0.001 \\
\textbackslash{} & 0.001 & 0.001 \\
\} & 0.001 & 0.001 \\
EOS & 0.000 & 0.000 \\
CLS & 0.000 & 0.000 \\ \bottomrule
\end{tabular}
\caption{Top-20 and bottom-20 tokens by average contribution to the similarity score in identical pairs. Values for level 8 of the \textit{Exact} model.}
\label{tab:lexical-attributions-exact}
\end{table}

\begin{table}[t]
\centering
\begin{tabular}{@{}lll@{}}
\toprule
\textbf{Word} & \textbf{Mean} & \textbf{StDev} \\ \midrule
auckland & 0.737 & 0.045 \\
cambodia & 0.713 & 0.098 \\
somme & 0.656 & 0.087 \\
sahara & 0.533 & 0.079 \\
shotgun & 0.514 & 0.032 \\
surgical & 0.507 & 0.127 \\
hacking & 0.503 & 0.143 \\
swiss & 0.502 & 0.116 \\
turkey & 0.496 & 0.150 \\
edmonton & 0.490 & 0.068 \\
anger & 0.477 & 0.093 \\
\#\#oop & 0.477 & 0.168 \\
pune & 0.472 & 0.124 \\
kerala & 0.461 & 0.084 \\
goa & 0.455 & 0.113 \\
coding & 0.455 & 0.169 \\
wikipedia & 0.454 & 0.116 \\
enfield & 0.453 & 0.114 \\
vocabulary & 0.449 & 0.086 \\
furniture & 0.447 & 0.103 \\ \midrule
their & 0.008 & 0.009 \\
’ & 0.007 & 0.006 \\
the & 0.007 & 0.007 \\
that & 0.007 & 0.007 \\
" & 0.005 & 0.009 \\
those & 0.005 & 0.011 \\
{[} & 0.004 & 0.010 \\
( & 0.004 & 0.010 \\
@ & 0.004 & 0.026 \\
 & 0.004 & 0.005 \\
, & 0.002 & 0.004 \\
ones & 0.002 & 0.004 \\
\{ & 0.002 & 0.004 \\
) & 0.002 & 0.006 \\
\textbackslash{} & 0.001 & 0.001 \\
\_ & 0.001 & 0.002 \\
{]} & 0.001 & 0.001 \\
\} & 0.000 & 0.000 \\
CLS & 0.000 & 0.001 \\
EOS & 0.000 & 0.000 \\ \bottomrule
\end{tabular}
\caption{Top-20 and bottom-20 tokens by average contribution to the similarity score in identical pairs. Values for level 8 of the \textit{Shelf} model.}
\label{tab:lexical-attributions-shelf}
\end{table}

\end{document}